# RTDK-BO: High Dimensional Bayesian Optimization with Reinforced Transformer Deep kernels

Alexander Shmakov, Avisek Naug, Vineet Gundecha, Sahand Ghorbanpour, Ricardo Luna Gutierrez,
Ashwin Ramesh Babu, Antonio Guillen and Soumyendu Sarkar*
Hewlett Packard Enterprise, USA
{alexander.shmakov, avisek.naug, vineet.gundecha, sahand.ghorbanpour, ricardo.luna,
ashwin.ramesh-babu, antonio.guillen, soumyendu.sarkar}@hpe.com,

*Abstract*— Bayesian Optimization (BO), guided by Gaussian process (GP) surrogates, has proven to be an invaluable technique for efficient, high-dimensional, black-box optimization, a critical problem inherent to many applications such as industrial design and scientific computing. Recent contributions have introduced reinforcement learning (RL) to improve the optimization performance on both single function optimization and *few-shot* multi-objective optimization. However, even few-shot techniques fail to exploit similarities shared between closely related objectives. In this paper, we combine recent developments in Deep Kernel Learning (DKL) and attention-based Transformer models to improve the modeling powers of GP surrogates with meta-learning. We propose a novel method for improving meta-learning BO surrogates by incorporating attention mechanisms into DKL, empowering the surrogates to adapt to contextual information gathered during the BO process. We combine this Transformer Deep Kernel with a learned acquisition function trained with continuous Soft Actor-Critic Reinforcement Learning to aid in exploration. This Reinforced Transformer Deep Kernel (RTDK-BO) approach yields state-of-the-art results in continuous high-dimensional optimization problems.

## I. INTRODUCTION

Efficient, high dimensional optimization is at the core of many industrial and scientific processes with applications including material design [1], physics [2], synthetic chemistry and biology [3], [4], and hyperparameter optimization [5]. In many cases, we are not just interested in optimizing a single situation or configuration, but rather many closely related problems, each with a slightly different configuration of a core process. For many industrial optimization problems, the evaluation of an objective function is performed through either costly simulations or scaled testing. Therefore, we must optimize these objectives with as few evaluations as possible, reducing the total cost required for running experiments.

Bayesian Optimization (BO) is a ubiquitous technique that has proven to be very promising across a variety of domains and is often the standard for sample-efficient black-box optimization. However, Bayesian Optimization relies on the design of a *surrogate model* which estimates the objective from observed samples, as well as an *acquisition function* for effectively exploring the optimization domain. Both of these components typically require domain knowledge to adapt to specific optimization objectives, and this design is critical to BO's performance on challenging domains. Additionally, BO typically optimizes a single objective at a time, reducing the ability to optimize many related objectives to reduce sample counts.

*corrosponding author: soumyendu.sarkar@hpe.com

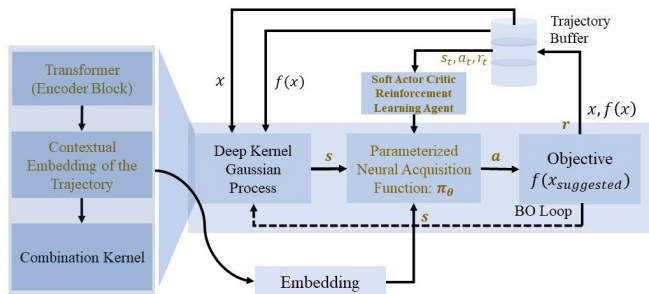

Fig. 1: RTDK-BO Architecture: Bayesian Optimization of a Target Function $f(X)$ using a shared embedding across the Deep Kernel Learning based GP and a Reinforcement Learning based acquisition function. This is with RL policy $\pi_1$, states s, action a, and reward r.

In this work, we aim to improve Bayesian Optimization's ability to tackle high dimensional optimization problems, using deep learning to both these components. We use contextual information (both x and y coordinates) in deep-learning models to learn a general embedding, which can help the Gaussian Process (GP) based surrogate model to effectively meta-learn across objectives from minimal function samples. In addition, we improve the Reinforcement Learning based acquisition function by reusing the learned embedding. We use transformers to generate the embedding and share it with both the Deep Kernel Learning based Gaussian model and the Reinforcement Learning based acquisition function. Our overall approach is summarized using Figure 1.

We evaluate our meta-learning optimization approach on both discrete and continuous optimization domains, focusing on families of high-dimensional continuous optimization objectives. We show a significant improvement in optimization of high dimensional objective functions, sometimes with added sample efficiency.

## II. RELATED WORKS

*a) High Dimensional Bayesian Optimization:* Bayesian Optimization has demonstrated great success in high dimen-

sion black box optimization [6], [7], proving very successful in many real world optimization tasks including hyperparameter search for machine learning. However, these methods typically focus on optimizing a single, relatively cheap, black-box function and may require thousands of evaluations. However, when limiting the function evaluation budget to the low hundreds, traditional BO suffers from an inability to meta-learn between problems.

*b) Reinforcement Learning Acquisitions:* Reinforcement learning has shown great success in applications such as healthcare, adversarial attacks, medical imaging, renewable energy, and many others [8]–[19]. Reinforcement learning (RL) approaches to Bayesian optimization have recently shown promising results, especially in meta-learning discrete objectives. MetaBO presents a seminal framework for interpreting the acquisition function within Bayesian optimization as a reinforcement learning policy, to be trained with policy gradient methods [20]. This work has been further generalized in [21] to allow for few-shot q-learning pre-trained on a set of randomly generated functions. These approaches present useful frameworks for tackling small-sample Bayesian optimization, but they primarily focus on the acquisition function, retaining the traditional Gaussian process surrogate model. Additionally, both approaches rely on a discrete grid of actions, approximating continuous domain optimization with a quasi-random hierarchical grid.

*c) Soft Actor-Critic:* Reinforcement learning in continuous action spaces presents several unique challenges compared to discrete action spaces such as exploration and policy representation. We will use soft actor-critic (SAC) [22], an off-policy RL algorithm, as our foundational algorithm and extend it to include the improvement from model-based methods and combine its actor, critic, and model to construct a robust acquisition function for Bayesian Optimization.

*d) Model-Based RL:* Model-based reinforcement learning presents a method for improving sample efficiency by estimating the environment's dynamics and allowing the policy to train from the information learned by such a model [23]. Additionally, it has been recently shown that the model must not necessarily estimate the exact dynamics, but can instead operate in a latent representation, allowing for a low dimensional representation [24]. We are interested in incorporating the sample-efficiency of model-based RL methods to improve black-box optimization methods.

*e) Attention and Transformers:* Attention mechanisms [25] provide a method for deep neural networks to modify their activations in response to a set of contextual vectors. Attention has been used in various network architectures to achieve state-of-the-art results in many applications, including natural language processing [26] and computer vision [27]. We will use the contextual embeddings of transformers to assist in rapidly meta-learning from observations originating from similarly structured objectives.

## III. BAYESIAN OPTIMIZATION

Bayesian Optimization (BO) defines a general set of techniques for optimizing a black-box function $f(x) : X \to Y$ by: fitting a *surrogate model* to estimate function values across the optimization domain, and employing an *acquisition function* to select promising query points [28].

The surrogate model, $\hat{f}(x; x_{train}, y_{train}) = \hat{f}(x; D)$, provides a probabilistic estimate of the objective across the entire optimization domain given a sparse sample of points $x_{train} = \{x_1, x_2, \ldots, x_k\}$ where the objective is known $y_{train} = \{y_1, y_2, \ldots, y_k\}$. Surrogates typically provide both a mean point estimate of the function value $\mu(x; D)$, as well as an uncertainty for that estimate in the form of a variance $\sigma^2(x; D)$.

After fitting the surrogate on the observed dataset D, the acquisition function, $A(x; D)$, defines a score estimating the promise of the next query point $x_{query}$. Bayesian optimization selects queries by maximizing the acquisition $x_{query} =_{x \in X} A(x; D)$. The acquisition is responsible for both balancing exploration to ensure a global optimum across the domain and exploitation to optimize locally within a promising region. One common acquisition function, especially with Gaussian Process surrogates, is the Expected Improvement (EI) criterion [29].

In this work, we aim to improve both aspects of BO by introducing deep neural networks to both the surrogate and acquisition functions while maintaining the generality of the BO approach. These improvements will minimize required training data while ensuring that these methods work in both continuous and discrete optimization domains.

## IV. CONDITIONAL DEEP KERNEL SURROGATE

The Gaussian Process (GP) [30] is a fundamental architecture for BO surrogate models [28]. GPs estimate the objective with a Gaussian posterior over functions fitting the observed data: $\hat{f}(y|x; D) \sim N(\mu(x; D), \sigma^2(x; D))$. We use a learned mean component, which is parameterized by a linear transformation of the input $\mu_W(x) = Wx$. In general, both the mean and covariance components may have parameters that must be learned. The parameters of a Gaussian process are trained by maximizing the conditional log-likelihood of the training dataset $\sum_{x,y \in D} \log P(\hat{f}(y; x, D) = y)$.

The GP covariance is determined by a *Kernel*, $K(x_1, x_2)$, defining a distance between any two points within a non-linear, high (possibly infinite) dimensional manifold. This kernel representation, along with the Gaussian likelihood, defines an analytical posterior distribution over the optimization domain given a set of observed function values. The choice of kernel function is crucial for well-fitting GPs, and many domains have specially designed kernel functions to fit domain-specific data.

### A. Combination Kernel

We wish to design a learnable kernel with the goal generality across various domains. A common kernel in traditional GPs is the RBF, which corresponds to a dot-product in an infinite dimensional manifold and may be defined as an infinite sum of polynomial kernels, expressed in Equation

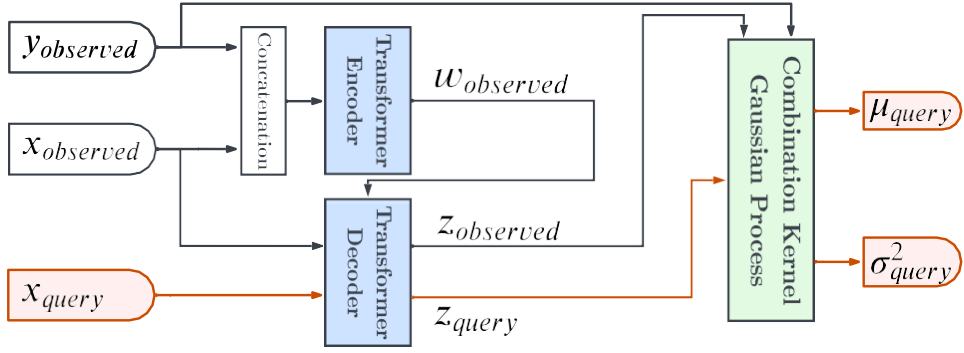

Fig. 2: Block Diagram of Transformer Deep Kernel Learning Gaussian Process. The red path indicates the prediction path for the query point. The black paths indicate the flow of contextual information.

1 [30].

$$K_{RBF}(x_1, x_2) \propto \sum_{k=1}^{\infty} \frac{\langle x_1, x_2 \rangle^k}{k!} \quad (1)$$

While there is a closed-form formulation to this kernel, in the interest of generalizing this basic GP, we extend the traditional RBF with additional learned parameters. Taking inspiration from this infinite sum representation, we choose to approximate a more general high dimensional embedding up to a certain power, $K$, learning the coefficients, $a_k$, as part of the kernel's parameters. We call this kernel the *Combination* up to $K$, or $C(K)$, kernel, defined in Equation 2.

$$K_{C(K)}(x_1, x_2) = \sum_{k=1}^{K} \frac{a_k \langle x_1, x_2 \rangle^k}{k!} \quad (2)$$

### B. Deep Kernel Learning

Deep Kernel Learning (DKL) [31], [32] extends the learning capability of the GP by mapping the original optimization domain $x \in X$ into a new latent domain via a parameterized transformation, such as a deep neural network $z = g_\theta(x) \in g_\theta(X)$. The underlying Gaussian process is then trained on these embedded inputs after neural network reprocessing: $\hat{f}(y|z; g(D)) \sim N(\mu(z), \sigma^2(z))$. The GP's parameters may be optimized via log-likelihood minimization, and the gradients can be passed along to the embedding $g_\theta$, optimizing both the GP and network parameters $\theta$ through back-propagation.

We find that DKL sometimes experiences numerical instability [32], especially when training for Bayesian Optimization tasks, which typically have very little data. We find that these instabilities largely arise from the off-diagonal covariance components. We, therefore, introduce an additional parameterized neural network, $v_\phi(x)$, to explicitly estimate the diagonal components of the covariance, while using a normalized variant [30] of the underlying kernel, $K(x_1, x_2)$, to estimate the off-diagonal components (Equation 3). This allows us to use a flexible, learnable kernel while avoiding large values due to a poorly conditioned embedding network.

$$K_{norm}(x_1, x_2) = \frac{\exp v_\phi(x_1) \exp v_\phi(x_2) K(x_1, x_2)}{K(x_1, x_1) K(x_2, x_2)} \quad (3)$$

### C. Transformer Deep Kernel Learning (TDKL)

Transformers present a mechanism for adding arbitrary context sequences to condition neural network activations. Crucially for BO, since the context does not necessarily require a unique target, it may include additional information that is not present in traditional BO observations. Specifically, we include not just the previous sample points $x_{obs}$ in the context, but also the known function values for those points $y_{obs}$. This mechanism has been shown to be sufficient for Bayesian inference by itself [33] and we follow a similar framework for conditioning a DKL embedding on previously observed data.

Formally, we extend the DKL framework to include a conditioning term on the embedding network, $z_{query} = g_\theta(x_{query}|x_{obs}, y_{obs})$, where $g$ is a sequence-to-sequence transformer encoder-decoder model [26]. The observed data, $(x_{obs}, y_{obs}) = \{(x_1, y_1), (x_2, y_2), \ldots, (x_K, y_K)\}$, is first fed through the transformer encoder to produce the latent encoded sequence $w_{obs} = \{w_1, w_2, \ldots, w_K\}$. This is used as the keys and values for the decoder, whereas the target sequence $x_{query} = \{x_{K+1}, x_{K+2}, \ldots, x_N\}$ is used as the query for the decoder. The output sequence, $z_{query} = \{z_{K+1}, z_{K+2}, \ldots, z_N\}$, represents a conditional embedding of the query locations. We also produce the conditional embedding of the original observed locations, $z_{obs} = \{z_1, z_2, \ldots, z_K\}$, to condition the downstream Gaussian Process. The output distribution is parameterized by our GP, $\hat{f}(y|z_{query}; z_{obs}, y_{obs})$. Figure 2 describes the flow diagram for all inputs. Similarly to [33], we remove the temporal embedding from the input to ensure that the transformer is invariant to sequence order. Additionally, we ensure that query points do not attend to each other by enforcing a diagonal attention mask on the decoder query sequence.

We refer to this complete surrogate model - employing a transformer embedding, learned point-wise variances, and a combination base kernel - as the Transformer Deep Kernel Learning (TDKL) surrogate. We extend the ideas introduced in [33] to focus on latent information in both the inputs and output function values from the history of optimization trajectories, while primarily relying on the Gaussian process mechanism to encode the uncertainty in our predictions.

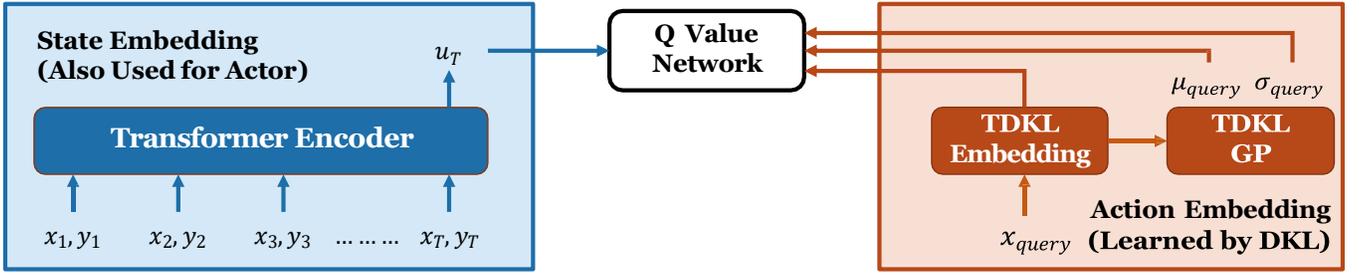

Fig. 3: Flow diagram for the critic network, showing how the state-action pair is embedded and processed into a Q-Value estimate.

Note that this design choice introduces an inductive bias towards smooth functions which may be represented by our combination kernel GP within the embedding space, but we hope that this assumption does limit our performance due to the nonlinear representation potential of the transformer.

## V. SOFT ACTOR CRITIC ACQUISITION

We turn our focus to the BO acquisition function. The TDKL described above provides a robust surrogate model which can estimate a function given a sufficiently robust set of samples. To collect such samples, we need a policy that will sufficiently explore the domain. For this purpose, we will employ a soft actor-critic (SAC) reinforcement learning agent [22]. We will represent the Bayesian optimization problem as a Markov decision process and use this formulation to train a novel model-based soft actor-critic agent.

### A. Optimization Environment

Following the formulation presented in MetaBO [20] and FSAF [21], the state space for the problem at any time $t$ of the optimization process is defined as the collection of points and values in the BO trajectory, $s = (x_{obs}, y_{obs}) = \{(x_1, y_1), (x_2, y_2), \ldots, (x_t, y_t)\}$. The action space corresponds to the space of possible locations where we can sample, defined by the input domain X, and action represents the next candidate point, $a_t = x_{t+1} \in X$. We define the reward in terms of the (approximate) regret, $r_{t+1} = -log(f^* - f(x_{t+1}))$, where $f^*$ is the estimate true optimum for the function. We consider finite-length trajectories determined by the budget for the number of steps in each trajectory $t < T_{max}$.

In general, each trajectory may originate from a different underlying objective. This enables meta-learning between similar objectives, as both the agent and model must perform well across many different trajectories. We may also collect additional trajectories from the same environment if we wish to continue optimizing the given objective. Sampled trajectories are stored in a replay buffer, separated by their objective variant.

### B. Model-Based Soft Actor-Critic

Following the soft actor-critic framework [22], our agent consists of two learned Q-value network $q_1(s, a)$ and $q_2(s, a)$, a conservative estimate of the q value $q(s, a) = \min\{q_1(s, a), q_2(s, a)\}$, and a probabilistic policy $\pi(s)$. We extend SAC to a model-based reinforcement learning method by introducing the learned surrogate model into the framework.

*a) Surrogate Model:* The surrogate model $\hat{f}$ will be trained on function samples from the replay buffer. After sampling a trajectory $s = \{(x_1, y_1), (x_2, y_2), \ldots, (x_T, y_T)\}$, we shuffle this trajectory and arbitrarily split the $x$ and $y$ values into *observation* and *query* datasets using a uniform random splitting pivot $M \sim U(1, T-1)$. These datasets $D_{obs} = \{(x_1, y_1), \ldots, (x_M, y_M)\}$ and $D_{query} = \{(x_{M+1}, y_{M+1}), \ldots, (x_T, y_T)\}$ are then used to optimize the surrogate on purely observed data.

$$L = \sum_{(x_{query}, y_{query}) \in D_{query}} L(x_{query}, y_{query})$$

$$= \sum_{(x_{query}, y_{query})} -\log P(\hat{f}(y; x_{query}, D_{obs}) = y_{query})$$

Notice that this loss function, Equation ??, differs from the traditional GP optimization because the training dataset which conditions our TDKL is different from the prediction dataset. This is to ensure that the TDKL learns a general representation regardless of which objective originated the trajectory. Optimizing this loss with a stochastic gradient-descent optimizer, sampling a different trajectory after each step, presents a cheap method for meta-learning GP models on a variety of objectives.

*b) Critics:* We train the dual critic networks using the standard entropy-corrected Bellman update described in [22]. However, we wish to include information learned by the surrogate and incorporate it in the critic networks to leverage the latent state space representation. We accomplish this by exploiting the learned embeddings of the TDKL and the GP estimates of the objective.

The state, $s$, is embedded using another transformer architecture [26]. This time, the transformer is acting only as an encoder to transform the variable-length observations, $s = \{(x_1, y_1), \ldots, (x_T, y_T)\}$, into a fixed-size latent embedding. We encode the sequence into a latent sequence using the transformer encoder $\{u_1^{critic}, \ldots, u_T^{critic}\} = TransformerEncoder(s)$ and then simply take the final latent vector, $u_T^{critic}$, as the fixed-length embedding.

The action, $a$, is encoded by passing the suggested point through the TDKL to extract both the embedding and function estimates from the surrogate model. $w_a = g_\theta(a|s))$ represents the embedding from the TDKL transformer $g_\theta$, and $\mu_a$, $\sigma_a^2 = \hat{f}(a|w_a; s)$ are the surrogate model estimates for the objective at the action location.

The Q-networks, therefore, become functions are the more abstract state-action representation, $q(u_T^{critic}, w_a, \mu_a, \sigma_a^2)$, allowing information to be shared between the model and critics. Note that TDKL parameters are treated as constant w.r.t the critic, and the gradient is not passed to the TDKL when optimizing the Bellman loss. A diagram of the critic architecture is presented in Figure 3.

*c) Actor:* The actor network, $\pi(s)$, uses the same architecture as the state encoding from the Q-networks (The left-hand component in Figure 3). We use a separately trained transformer encoder to construct a fixed-length embedding for the state representation. $u_T^{actor} \in TransformerEncoder(s)$. Following the methods described in [22], we use a tanh-squashed normal for the actor distribution and apply an affine transform to fit the desired optimization domain.

### C. Actions in Continuous Domains with Importance Sampling

We wish to generalize our approach so that it works also in domains where sample evaluations are expensive. We found that relying purely on the actor for selecting good actions while exploring performs poorly on the small sample counts found in Bayesian optimization. Therefore, instead of sampling the action directly from $a \sim \pi(s)$ like in traditional SAC, we would like to incorporate the Q networks into the inference policy.

To do this, we take inspiration from Boltzmann exploration [34], a common exploration technique in discrete environments. This involves constructing a policy that will sample proportional to a Boltzmann distribution based on the Q-network, $P(a|s) \propto \exp Q(s, a)$. We perform this kind of Boltzmann sampling when optimizing discrete domains using the Softmax function, but this approach cannot be generalized to continuous optimization domains.

Generally, sampling from an arbitrary continuous domain can be difficult, especially in high dimensions. However, if we believe that our actor approximates the landscape of our Q-function, then we can sample from the actor in order to assist with generating samples from the Boltzmann Q using **importance sampling**.

First, we sample a large batch of actions from the policy, $\{a_1, a_2, \ldots, a_N\} \sim \pi(s)$, where $N$ is typically in the thousands. Then, we compute the importance weights of each sampled points w.r.t the Boltzmann Q, $w_i = \frac{\exp(Q(s,a_i))}{P(\pi(s)=a_i)}$. Finally, we construct an empirical distribution over $\{a_1, a_2, \ldots, a_N\}$ using these importance weights and sample an action $a$ such that $P(a = a_i) = \frac{w_i}{\sum_{k=1}^{N} w_k}$. If the actor has a non-zero probability of sampling anywhere in the optimization domain, this process will produce samples from exactly the Boltzmann Q distribution as $N \rightarrow \infty$ [35]. In practice, as a trade-off between memory and computation time, we use a value of $N = 1024$.

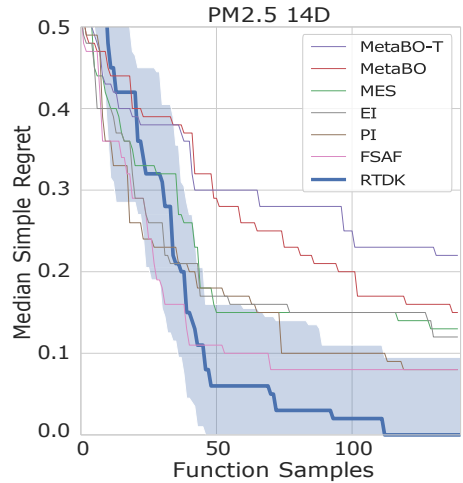

Fig. 4: A comparison of median achieved regret value for the discrete domain optimization objective PM 2.5 14-dimensional. Each run contains 36 unique variations. RTDK-BO results include the shaded region representing the IQR (25% - 75%) of achieved regret value.

## VI. EXPERIMENTS

We evaluate the RTDK-BO Bayesian optimization approach on a variety of test functions. First, we confirm that this fully optimizable approach can still achieve reliable results in simpler discrete optimization tasks while comparing common baselines used by other RL optimization approaches. Then we explore the effectiveness of this fully continuous approach on high dimensional optimization problems.

The functions we present are well-understood baseline test functions and are not expensive to evaluate. We perform experiments on these functions because we may always compute the global optimum using non-sample-efficient methods and evaluate the performance of the tested algorithms with a known optimal baseline. Since our methods focus on high dimensions, high-cost black-box optimization, we will be operating under an artificial restriction that we are only allowed to sample a small number of function samples for each optimization run. We, therefore, limit every experiment to only 150 function samples for each objective and 250 total function samples for pre-training and meta-learning. This will allow us to explore these methods in extremely low-sample conditions.

We evaluate the RTDK-BO with a transformer DKL embedding network and a combination base kernel with 5 components $C(5)$. RTDK-BO evaluation differs slightly from other BO methods because the transformer models must be trained on entire trajectories. Therefore, we must present the model with similarly sized trajectories during both training and evaluation. To accommodate longer evaluation trajectories, we split the BO run into "sub-trajectories" of length 50, resetting the surrogate after 50 steps. This allows us to train on trajectories up to length 50, reducing over-fitting by re-sampling longer trajectories. Each trajectory also contains 5 uniform random function samples to initialize the RTDK-BO surrogate. Additionally, we add a large action exploration to the first sub-trajectory in order to initialize the RTDK-BO with a well-conditioned set of points. We examine the effect of these sub-trajectories in Figure 6.

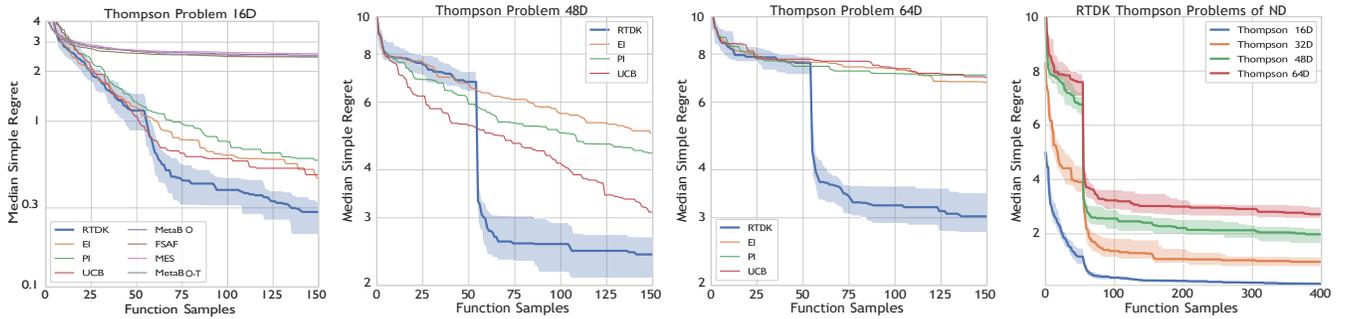

Fig. 5: A comparison of different optimization methods on high dimensional continuous optimization tasks. The final plot shows a comparison of RTDK-BO across different dimensional Thompson optimization runs.

*A. Discrete Optimization Baseline*

We evaluate RTDK-BO, as well as a variety of baselines, on discrete, real-world optimization tasks. In order to aid in direct comparison, we evaluate this method on one of the most difficult high-dimensional problems mentioned in [21] i.e. the PM 2.5 14D problem.

We compare against baseline acquisitions based on a traditional Gaussian process including Expected Improvement (EI) [36], Probability of Improvement (PI) [37], and Max-value entropy search (MES) [38]. We also compare against the current state-of-the-art in deep learning acquisition functions: FSAF [21] and MetaBO [20] and its variations to compare against other meta-learning techniques. We allow the models which can perform meta-learning to pre-train on 250 function samples: 5 trajectories of 50 samples each. We then evaluated all the methods on 36 additional variants, and plot the median regret values for these evaluation runs in Figure 4.

RTDK-BO had to be adapted to perform discrete optimization by limiting the acquisition function to only valid discrete locations. The SAC RL agent was still trained as a continuous acquisition function. This presents an opportunity to improve this method in the future, since there is a specialization of SAC to discrete action spaces [39]. However, we wanted to keep the method consistent across all experiments and employed the same continuous SAC for all cases.

We find from figure 4 that RTDK-BO clearly outperforms both traditional and reinforcement learning based acquisition functions on the high dimensional PM2.5 data set. Crucially, as dimensionality increases, we find that the transformer surrogate starts to help optimize the objective better when compared to the other two RL-enhanced methods, MetaBO and FSAF as we shall see in the next results.

*B. High Dimensional Continuous Domains*

Comparison of DKL architecture in meta-learning optimization (Thompson 16D).

We perform a similar comparison of optimization methods in the continuous domain. For these experiments, we evaluate high dimensional variations of the Thompson Problem [40]. This optimization objective consists of placing $N$ electrons on the surface of a unit sphere, with the objective of minimizing the potential energy between all pairs of electrons. We parameterize this function in $4N$ dimensions, storing the *sin* and *cos* of both the polar and azimuthal angle of each electron on the sphere. We construct the 16, 32, 48, and 64 dimensional meta-learnable variants of the Thompson problem by taking random $D$-dimensional slices through a $N = 32$ Thompson problem. Each slice is treated as a variant of the function, allowing us to make a near-infinite amount of objectives with similar characteristics.

For this test, we continue using the discrete grid-based approach described in [20], [21] for baseline methods FSAF, MetaBO and MES. This is because generalizing these methods to the continuous domain directly is challenging, and [21] recommends using a discrete grid of quasi-random Sobol samples for approximating continuous optimization. However, other baseline methods - EI, PI, and UCB - may be adapted for use with continuous Gaussian processes and optimization schemes. Therefore, we use a continuous Bayesian optimization routine for these baselines, along with the RTDK-BO model.-BO

The combined benefits of the transformer surrogate and SAC acquisition outperform the other methods on the higher dimensional continuous optimization problems presented in Figure 5. We found that the discrete methods failed to optimize within these high dimensional domains due to the exponential growth in the required grid size to densely cover the domain. This presents a key improvement for this technique among RL-enhanced optimization methods. We also find that RTDK-BO starts t-BOo break off from the baseline methods and effectively meta-learns on the 48 and 64-dimensional Thompson problems. We also observe a discrete phase transition between the first 50-sample sub-trajectory and the later sub-trajectories for the RTDK-BO model. Due to the exploration annealing and buffered nature of transformers, the first 50 samples effectively serve as a "warm-up" phase where the IE exploration is collecting information. Once RTDK-BO receives enough data to form a full sub-trajectory, it rapidly jumps in performance after the TDKL surrogate initializes. We plan to look into this behavior to allow for a smoother transition, potentially improving sample efficiency.

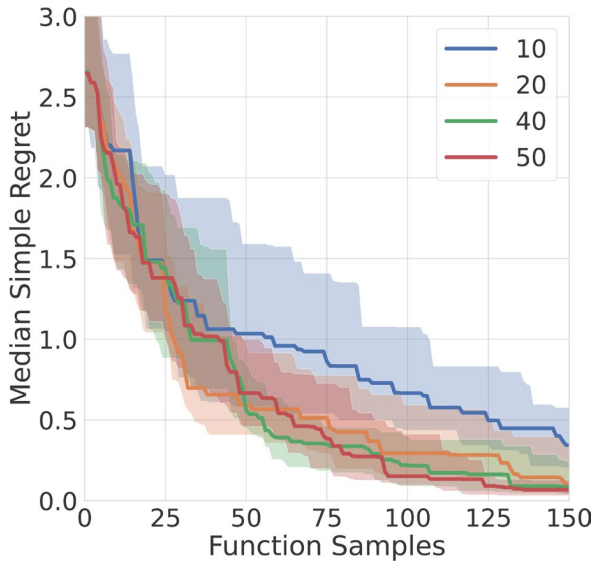 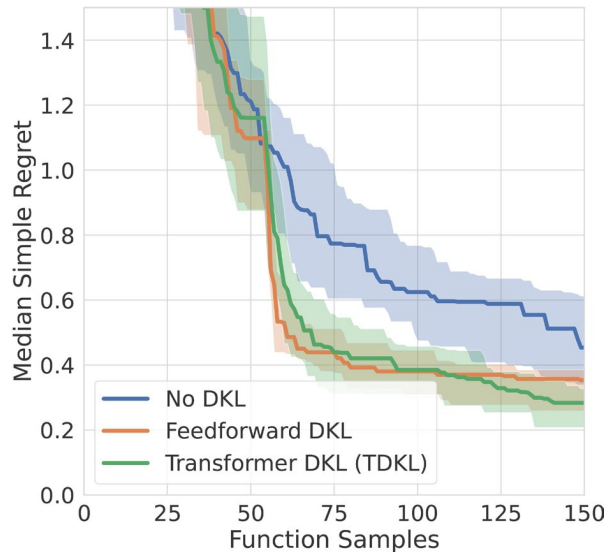

Fig. 6: Ablation studies on RTDK-BO model, enabling and disabling various components. Left: Comparison of sub-trajectory length on regret in longer episodes. Right: Comparison of DKL architecture in meta-learning optimization (Thompson 16D).

*C. Ablation Study*

*a) Sub-Trajectory Length:* We find that longer sub-trajectories assist with achieving lower final regret on the 10-Dimensional Powell function (Figure 6a). However, this comes with a slight hit to convergence time, taking longer to achieve low regret earlier during the optimization. Additionally, sub-trajectories larger than 50 samples risked running out of memory as the transformer architecture memory requirements scale as $n^2$ with respect to the sequence length. However, we find that performance plateaus after length 30, so these longer trajectories are not necessary for optimization. For the experiments, we used a trajectory length of 50 to ensure the lowest possible regret with reasonable memory usage. However, in practice, lower lengths may be useful to improve performance time and improve sample efficiency in very sample-limited situations by avoiding the "warm-up" effect seen with the Thompson problem.

*b) Meta-Learning DKL Architecture:* In Figure 6b, we evaluate the no DKL and DKL structure variants (feedforward and transformer) on optimizing the 16-dimensional Thompson problem. We see clear separation when evaluating the three different DKL approaches on the meta-learning task. This objective has more meta-learning because each objective is a projection of a 128-dimensional Thompson problem projected onto a random 96-dimensional plane. This results in a wide variety of possible objective landscapes. We find that the Transformer DKL achieves lower final regret values when compared to the feed-forward approach.

## VII. CONCLUSION

We introduced Reinforced Transformer Deep Kernel (RTDK-BO), a novel approach for continuous high dimensional Bayesian Optimization. It uses a Soft Actor-Critic based neural acquisition function that learns a contextual embedding from optimization trajectories using Transformer Encoders. The proposed method can meta-learn from past trajectories and use it to embed context during searching for candidate locations using the RL-based neural acquisition function. We experimentally showed that RTDK-BO performs competitively against traditional acquisition functions like Expected Improvement, Probability of Improvement as well as meta-learning Deep RL acquisition functions such as MetaBO and FSAF by obtaining lower regret values consistently. While RTDK-BO performs well on high-dimensional continuous control tasks, we plan to expand its ability to perform well on discrete optimization tasks in the future. Furthermore, we plan to improve our approach on low dimensional optimization tasks so that it can work in a sample efficient manner compared to classical approaches.

REFERENCES

[1] Yichi Zhang, Daniel W. Apley, and Wei Chen. Bayesian optimization for materials design with mixed quantitative and qualitative variables. *Scientific Reports*, 10(1):4924, Mar 2020.
[2] Shane Carr, Roman Garnett, and Cynthia Lo. Basc: Applying bayesian optimization to the search for global minima on potential energy surfaces. In Maria Florina Balcan and Kilian Q. Weinberger, editors, *Proceedings of The 33rd International Conference on Machine Learning*, volume 48 of *Proceedings of Machine Learning Research*, pages 898–907, New York, New York, USA, 20–22 Jun 2016. PMLR.
[3] Benjamin J. Shields, Jason Stevens, Jun Li, Marvin Parasram, Farhan Damani, Jesus I. Alvarado, Jacob M. Janey, Ryan P. Adams, and Abigail G. Doyle. Bayesian reaction optimization as a tool for chemical synthesis. *Nature*, 590(7844).
[4] Chris P. Barnes, Daniel Silk, Xia Sheng, and Michael P. H. Stumpf. Bayesian design of synthetic biological systems. *Proceedings of the National Academy of Sciences*, 108(37):15190–15195, 2011.
[5] Jasper Snoek, Hugo Larochelle, and Ryan P Adams. Practical bayesian optimization of machine learning algorithms. In F. Pereira, C.J. Burges, L. Bottou, and K.Q. Weinberger, editors, *Advances in Neural Information Processing Systems*, volume 25. Curran Associates, Inc., 2012.

[6] Zi Wang, Clement Gehring, Pushmeet Kohli, and Stefanie Jegelka. Batched large-scale bayesian optimization in high-dimensional spaces. In Amos Storkey and Fernando Perez-Cruz, editors, *Proceedings of the Twenty-First International Conference on Artificial Intelligence and Statistics*, volume 84 of *Proceedings of Machine Learning Research*, pages 745–754. PMLR, 09–11 Apr 2018.

[7] David Eriksson and Martin Jankowiak. High-dimensional Bayesian optimization with sparse axis-aligned subspaces. In Cassio de Campos and Marloes H. Maathuis, editors, *Proceedings of the Thirty-Seventh Conference on Uncertainty in Artificial Intelligence*, volume 161 of *Proceedings of Machine Learning Research*, pages 493–503. PMLR, 27–30 Jul 2021.

[8] Soumyendu Sarkar, Ashwin Ramesh Babu, Sajad Mousavi, Sahand Ghorbanpour, Vineet Gundecha, Ricardo Luna Gutierrez, Antonio Guillen, and Avisek Naug. Reinforcement learning based black-box adversarial attack for robustness improvement. In *2023 IEEE 19th International Conference on Automation Science and Engineering (CASE)*, pages 1–8. IEEE, 2023.

[9] Soumyendu Sarkar, Ashwin Ramesh Babu, Sajad Mousavi, Vineet Gundecha, Sahand Ghorbanpour, Alexander Shmakov, Ricardo Luna Gutierrez, Antonio Guillen, and Avisek Naug. Robustness with blackbox adversarial attack using reinforcement learning. In *AAAI 2023: Proceedings of the Workshop on Artificial Intelligence Safety 2023 (SafeAI 2023)*, volume 3381. https://ceur-ws.org/Vol-3381/8.pdf, 2023.

[10] Soumyendu Sarkar, Ashwin Ramesh Babu, Vineet Gundecha, Antonio Guillen, Sajad Mousavi, Ricardo Luna, Sahand Ghorbanpour, and Avisek Naug. Robustness with query-efficient adversarial attack using reinforcement learning. In *Proceedings of the IEEE/CVF Conference on Computer Vision and Pattern Recognition*, pages 2329–2336, 2023.

[11] Soumyendu Sarkar, Avisek Naug, Antonio Guillen, Ricardo Luna Gutierrez, Sahand Ghorbanpour, Sajad Mousavi, Ashwin Ramesh Babu, and Vineet Gundecha. Concurrent carbon footprint reduction (c2fr) reinforcement learning approach for sustainable data center digital twin. In *2023 IEEE 19th International Conference on Automation Science and Engineering (CASE)*, pages 1–8, 2023.

[12] Rahman Ejaz, Varchas Gopalaswamy, Aarne Lees, Duc Cao, Soumyendu Sarkar, and Christopher Kanan. Direct-drive implosion performance optimization using gaussian process modeling and reinforcement learning. *Bulletin of the American Physical Society*, 2023.

[13] Soumyendu Sarkar, Ashwin Ramesh Babu, Vineet Gundecha, Antonio Guillen, Sajad Mousavi, Ricardo Luna, Sahand Ghorbanpour, and Avisek Naug. Rl-cam: Visual explanations for convolutional networks using reinforcement learning. In *Proceedings of the IEEE/CVF Conference on Computer Vision and Pattern Recognition*, pages 3860–3868, 2023.

[14] Soumyendu Sarkar, Sajad Mousavi, Ashwin Ramesh Babu, Vineet Gundecha, Sahand Ghorbanpour, and Alexander K Shmakov. Measuring robustness with black-box adversarial attack using reinforcement learning. In *NeurIPS ML Safety Workshop*, 2022.

[15] Soumyendu Sarkar, Vineet Gundecha, Sahand Ghorbanpour, Alexander Shmakov, Ashwin Ramesh Babu, Alexandre Pichard, and Mathieu Cocho. Skip training for multi-agent reinforcement learning controller for industrial wave energy converters. In *2022 IEEE 18th International Conference on Automation Science and Engineering (CASE)*, pages 212–219. IEEE, 2022.

[16] Soumyendu Sarkar, Vineet Gundecha, Alexander Shmakov, Sahand Ghorbanpour, Ashwin Ramesh Babu, Paolo Faraboschi, Mathieu Cocho, Alexandre Pichard, and Jonathan Fievez. Multi-objective reinforcement learning controller for multi-generator industrial wave energy converter. In *NeurIPs Tackling Climate Change with Machine Learning Workshop*, 2021.

[17] Soumyendu Sarkar, Vineet Gundecha, Alexander Shmakov, Sahand Ghorbanpour, Ashwin Ramesh Babu, Paolo Faraboschi, Mathieu Cocho, Alexandre Pichard, and Jonathan Fievez. Multi-agent reinforcement learning controller to maximize energy efficiency for multigenerator industrial wave energy converter. In *Proceedings of the AAAI Conference on Artificial Intelligence*, volume 36, pages 12135–12144, 2022.

[18] Sajad Mousavi, Ricardo Luna Gutiérrez, Desik Rengarajan, Vineet Gundecha, Ashwin Ramesh Babu, Avisek Naug, Antonio Guillen, and Soumyendu Sarkar. N-critics: Self-refinement of large language models with ensemble of critics, 2023.

[19] Soumyendu Sarkar, Vineet Gundecha, Sahand Ghorbanpour, Alexander Shmakov, Ashwin Ramesh Babu, Avisek Naug, Alexandre Pichard, and Mathieu Cocho. Function approximation for reinforcement learning controller for energy from spread waves. In Edith Elkind, editor, *Proceedings of the Thirty-Second International Joint Conference on Artificial Intelligence, IJCAI-23*, pages 6201–6209. International Joint Conferences on Artificial Intelligence Organization, 8 2023. AI for Good.

[20] Michael Volpp, Lukas P. Fröhlich, Kirsten Fischer, Andreas Doerr, Stefan Falkner, Frank Hutter, and Christian Daniel. Meta-learning acquisition functions for transfer learning in bayesian optimization. In *International Conference on Learning Representations*, 2020.

[21] Bing-Jing Hsieh, Ping-Chun Hsieh, and Xi Liu. Reinforced few-shot acquisition function learning for bayesian optimization. In *Conference on Neural Information Processing Systems*, 2021.

[22] Tuomas Haarnoja, Aurick Zhou, Pieter Abbeel, and Sergey Levine. Soft actor-critic: Off-policy maximum entropy deep reinforcement learning with a stochastic actor. In Jennifer Dy and Andreas Krause, editors, *Proceedings of the 35th International Conference on Machine Learning*, volume 80 of *Proceedings of Machine Learning Research*, pages 1861–1870. PMLR, 10–15 Jul 2018.

[23] Richard S. Sutton. Dyna, an integrated architecture for learning, planning, and reacting. *SIGART Bull.*, 2(4):160–163, jul 1991.

[24] Junhyuk Oh, Satinder Singh, and Honglak Lee. Value prediction network. *CoRR*, abs/1707.03497, 2017.

[25] Thang Luong, Hieu Pham, and Christopher D. Manning. Effective approaches to attention-based neural machine translation. In *Proceedings of the 2015 Conference on Empirical Methods in Natural Language Processing*, pages 1412–1421, Lisbon, Portugal, September 2015. Association for Computational Linguistics.

[26] Ashish Vaswani, Noam Shazeer, Niki Parmar, Jakob Uszkoreit, Llion Jones, Aidan N Gomez, Ł ukasz Kaiser, and Illia Polosukhin. Attention is all you need. In I. Guyon, U. Von Luxburg, S. Bengio, H. Wallach, R. Fergus, S. Vishwanathan, and R. Garnett, editors, *Advances in Neural Information Processing Systems*, volume 30. Curran Associates, Inc., 2017.

[27] Alexey Dosovitskiy, Lucas Beyer, Alexander Kolesnikov, Dirk Weissenborn, Xiaohua Zhai, Thomas Unterthiner, Mostafa Dehghani, Matthias Minderer, Georg Heigold, Sylvain Gelly, Jakob Uszkoreit, and Neil Houlsby. An image is worth 16x16 words: Transformers for image recognition at scale. In *International Conference on Learning Representations*, 2021.

[28] Peter I. Frazier. A tutorial on bayesian optimization, 2018.

[29] Donald R. Jones, Matthias Schonlau, and William J. Welch. Efficient global optimization of expensive black-box functions. *Journal of Global Optimization*, 13(4):455–492, Dec 1998.

[30] Carl Edward Rasmussen and Christopher K. I. Williams. *Gaussian Processes for Machine Learning*. The MIT Press, 11 2005.

[31] Andrew Gordon Wilson, Zhiting Hu, Ruslan Salakhutdinov, and Eric P. Xing. Deep kernel learning. *CoRR*, abs/1511.02222, 2015.

[32] Sebastian W. Ober, Carl E. Rasmussen, and Mark van der Wilk. The Promises and Pitfalls of Deep Kernel Learning. *arXiv e-prints*, page arXiv:2102.12108, February 2021.

[33] Samuel Müller, Noah Hollmann, Sebastian Pineda Arango, Josif Grabocka, and Frank Hutter. Transformers can do bayesian-inference by meta-learning on prior-data. In *Fifth Workshop on Meta-Learning at the Conference on Neural Information Processing Systems*, 2021.

[34] Nicolò Cesa-Bianchi, Claudio Gentile, Gábor Lugosi, and Gergely Neu. Boltzmann exploration done right. In *Proceedings of the 31st International Conference on Neural Information Processing Systems*, NIPS'17, page 6287–6296, Red Hook, NY, USA, 2017. Curran Associates Inc.

[35] T. Kloek and H. K. van Dijk. Bayesian estimates of equation system parameters: An application of integration by monte carlo. *Econometrica*, 46(1):1–19, 1978.

[36] J. Močkus. On bayesian methods for seeking the extremum. In G. I. Marchuk, editor, *Optimization Techniques IFIP Technical Conference Novosibirsk, July 1–7, 1974*, pages 400–404, Berlin, Heidelberg, 1975. Springer Berlin Heidelberg.

[37] H. J. Kushner. A New Method of Locating the Maximum Point of an Arbitrary Multipeak Curve in the Presence of Noise. *Journal of Basic Engineering*, 86(1):97–106, 03 1964.

[38] Zi Wang and Stefanie Jegelka. Max-value entropy search for efficient Bayesian optimization. In Doina Precup and Yee Whye Teh, editors, *Proceedings of the 34th International Conference on Machine Learning*, volume 70 of *Proceedings of Machine Learning Research*, pages 3627–3635. PMLR, 06–11 Aug 2017.

[39] Petros Christodoulou. Soft actor-critic for discrete action settings. *CoRR*, abs/1910.07207, 2019.

[40] J.J. Thomson F.R.S. Xxiv. on the structure of the atom: an investigation of the stability and periods of oscillation of a number of corpuscles arranged at equal intervals around the circumference of a circle; with application of the results to the theory of atomic structure. *The London, Edinburgh, and Dublin Philosophical Magazine and Journal of Science*, 7(39):237–265, 1904.